\documentclass[10pt,twocolumn,letterpaper]{article}

\usepackage{iccv}
\usepackage{times}
\usepackage{epsfig}
\usepackage{graphicx}
\usepackage{amsmath}
\usepackage{amssymb}
\usepackage{multirow}
\usepackage{xcolor}         
\usepackage{algorithm}
\usepackage{enumitem}

\usepackage{algpseudocode}
\usepackage{amsfonts}

\usepackage[pagebackref=true,breaklinks=true,letterpaper=true,colorlinks,bookmarks=false]{hyperref}
\usepackage{cleveref}

\iccvfinalcopy 

\def\iccvPaperID{12695} 

\def\Modelname{\textsc{UCoFiA}}

\begin{document}

\title{Unified Coarse-to-Fine Alignment for Video-Text Retrieval}

\author{Ziyang Wang, Yi-Lin Sung, Feng Cheng, Gedas Bertasius, Mohit Bansal\\
UNC Chapel Hill\\
{\tt\small \{ziyangw, ylsung, fengchan, gedas, mbansal\}.cs.unc.edu}
}

\maketitle

\begin{abstract}
The canonical approach to video-text retrieval leverages a coarse-grained or fine-grained alignment between visual and textual information. However, retrieving the correct video according to the text query is often challenging as it requires the ability to reason about both high-level (scene) and low-level (object) visual clues and how they relate to the text query. To this end, we propose a \textbf{U}nified \textbf{Co}arse-to-\textbf{fi}ne \textbf{A}lignment model, dubbed \Modelname{}. Specifically, our model captures the cross-modal similarity information at different granularity levels. To alleviate the effect of irrelevant visual clues, we also apply an Interactive Similarity Aggregation module (ISA) to consider the importance of different visual features while aggregating the cross-modal similarity to obtain a similarity score for each granularity. Finally, we apply the Sinkhorn-Knopp algorithm to normalize the similarities of each level before summing them, alleviating over- and under-representation issues at different levels. By jointly considering the cross-modal similarity of different granularity, \Modelname{} allows the effective unification of multi-grained alignments. Empirically, \Modelname{} outperforms previous state-of-the-art CLIP-based methods on multiple video-text retrieval benchmarks, achieving $2.4\%$, $1.4\%$ and $1.3\%$ improvements in text-to-video retrieval R@1 on MSR-VTT, Activity-Net, and DiDeMo, respectively. Our code is publicly available at \url{https://github.com/Ziyang412/UCoFiA}.

\end{abstract}

\section{Introduction}
The fields of computer vision and natural language processing have both seen significant progress in recent years. Thus, the cross-modal alignment \cite{alayrac2022flamingo, zellers2022merlot, lei2021less, radford2021learning, yu2023self, wu2021star, sung2022vl, sung2022lst, sung2023empirical}, which involve developing techniques to connect these two domains, has seen considerable attention and progress. As a direct application of cross-modal alignment, the video-text retrieval task aligns video(text) candidates with text(video) queries to identify the most relevant videos, and the standard practice \cite{torabi2016learning, yu2018joint, yu2016video} is to align the video and text features extracted by vision and language encoders. Recently, the emergence of large-scale image-text pretrained models prompted several methods \cite{luo2022clip4clip, ma2022x, liu2022ts2} to utilize CLIP~\cite{radford2019language} image and text encoder to achieve strong performance on many video-text retrieval benchmarks. As a direct extension of CLIP, Luo \etal~ \cite{luo2022clip4clip} proposed temporal fusion modules to aggregate the features of different video frames and then perform the cross-modal alignment on video and text features. Later, to capture more correspondences between video and text, several works \cite{liu2022ts2, gorti2022x, fang2021clip2video, cheng2021improving} propose to conduct the alignment between frame and text features. Ma \etal~\cite{ma2022x} take a step forward and leverage an alignment between frame and word features for more detailed information.  

\begin{figure}[t]
\begin{center}
   \includegraphics[width=1\linewidth]{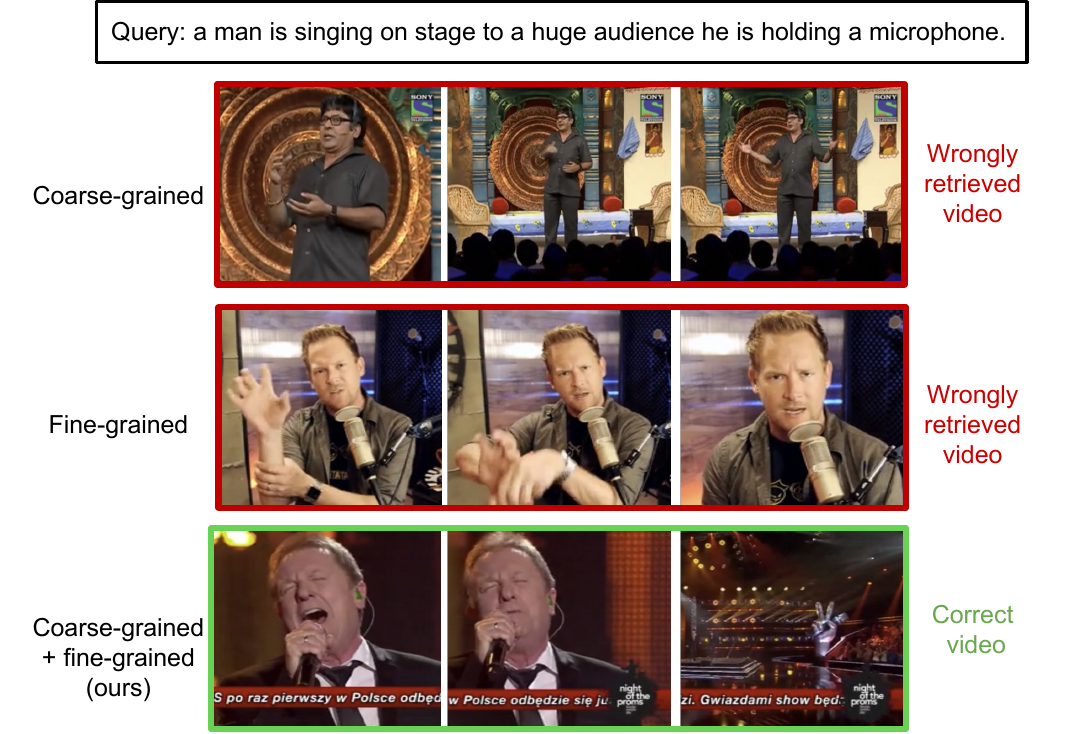}
\end{center}
   \caption{Comparison of the retrieved video from different level  cross-modal alignments given a specific query. The coarse-grained alignment (the first row) only observes the high-level scene information and overlooks the detailed information like ``microphone''. The fine-grained alignment (the second row) does capture the detailed information but ignores the high-level scene information like ``stage with the huge audience''. To this end, our work (the last row) combines the strengths of coarse-grained and fine-grained alignments by a Unified Coarse-to-fine Alignment model (\Modelname{}). }
\label{fig:intro}
\label{fig:movitation}
\end{figure}

Although the aforementioned methods have achieved impressive results, they only rely on high-level visual information (frame and video) to perform the cross-modal alignment. This coarse-grained alignment only captures the high-level visual clues (scene, action, etc) that connect to the text query. As shown in the first row of \Cref{fig:intro}, the coarse-grained alignment only captures the scene of the stage with the huge audience'' and the action of ``singing (possibly talking)'', thus leading to the incorrect retrieval result. On the other hand, Zou \etal~\cite{zou2022tokenflow} build a fine-grained alignment between patch tokens from the video and word tokens from the text query. As illustrated in the second row of \Cref{fig:intro}, the fine-grained alignment does capture the detailed information like ``microphone'', but it might overlook high-level clues like scene information (``stage with the huge audience''). These results reveal that video-text retrieval requires an understanding of the both high-level and low-level correspondence between text and video. Thus, in this work, we aim to jointly consider coarse-grained and fine-grained cross-modal alignment and how to unify them to get the correct answer (as shown in the last row of \Cref{fig:intro}).

To this end, we propose \Modelname{}, a \textbf{U}nified \textbf{Co}arse-to-\textbf{fi}ne \textbf{A}lignment model for video-text retrieval. 
Our approach aims to capture the multi-grained similarity between the text and video by performing alignment at different granularity. We begin with a coarse-grained alignment between the entire video and the query sentence (video-sentence). Next, we perform frame-sentence alignment by matching individual video frames and the query sentence. Finally, we conduct a fine-grained alignment between the video patches and query words (patch-word).

However, while this multi-grained information provides richer, more diverse detailed information, it also brings significant irrelevant information to the cross-modal alignment. For instance, several frames in the video might not contain information related to the query, and some patches in a frame might only correspond to the background information unrelated to any subjects in the query. The irrelevant information could impede the model from learning precise cross-modal correspondence. To address these issues, we first propose an Interactive Similarity Aggregation module (ISA) that considers the importance of different visual features while aggregating the cross-modal similarity to obtain a similarity score for each granularity. For frame-sentence alignment, our ISA module jointly considers the cross-modal similarity and the interaction of frame features while aggregating the frame-sentence similarity. Compared to the previous methods \cite{liu2022ts2} that ignore the temporal clues between video frames, our ISA module can better capture the important information within continuous video frames. Note that the ISA module is a general similarity aggregation approach regardless of the feature granularity, and we further extend it to a bidirectional ISA module for patch-word alignment. 

Next, once we obtain the similarity score for each level of alignment, we can sum them to one score as the final retrieval similarity. However, we find that similarity scores across different videos are highly imbalanced, and we empirically show that correcting this imbalance before summation improves the performance. Concretely, sometimes the sum of retrieval similarities between one specific video and all texts (we called this term marginal similarity) might be much higher than that of the other videos, meaning that this video is over-represented and will lower the probability of the other video being selected. To address this, we utilize the Sinkhorn-Knopp algorithm \cite{cuturi2013sinkhorn} to normalize the similarity scores and make sure the marginal similarities for different videos are almost identical so that each video has a fair chance to be selected after normalization. We then unify the scores of different levels by performing the algorithm separately on the similarities of different levels and summing them together.

We validate the effectiveness of our \Modelname{} model on diverse video-text retrieval benchmarks. Specifically, \Modelname{} achieve a text-to-video retrieval R@1 of $49.4\%$ on MSR-VTT \cite{xu2016msr} and $45.7\%$ on ActivityNet, thus, outperforming the current state-of-the-art CLIP-based methods by $2.4\%$ and $1.4\%$, respectively.

\section{Related Work}
\noindent  \textbf{Video-text Retrieval}. 
Video-text retrieval \cite{yu2018joint, lin2023vision, croitoru2021teachtext, yang2021taco, wang2021t2vlad, lin2022eclipse, chen2020uniter, alayrac2022flamingo, zellers2022merlot,chen2020fine, wang2023video, lu2023uniadapter, cheng2023cico, shu2022masked} is a fundamental topic in the vision-language domain and has attracted significant research attention. To retrieve the correct video candidate given the text query, it is crucial to align the features of the related video and text sample together. To this end, early works in video-text retrieval \cite{torabi2016learning, yu2018joint, yu2016video} focus on designing fusion mechanisms for the alignment between pre-extracted and frozen video and text features. Later, ClipBERT \cite{lei2021less} proposes a sparse sampling strategy on video data to accomplish end-to-end training and apply image-text pretraining for video-text retrieval. Afterward, Bain \etal~\cite{bain2021frozen} utilize a curriculum learning schedule to accomplish joint image and video end-to-end training on cross-modal data. With the great success of large-scale image-text pretraining model CLIP \cite{radford2021learning}, several works \cite{luo2022clip4clip, bain2022clip, fang2021clip2video, buch2022revisiting, jiang2022cross} utilize the powerful CLIP encoder for video-text retrieval tasks and achieve state-of-the-art results with an efficient training paradigm. Thus, in this work, we also use CLIP as our image-text backbone to enable a fair comparison with existing methods. 

Moreover, most cross-modal alignment approaches can be divided into two categories: coarse-grained alignment \cite{li2021align, dou2022empirical, jia2021scaling, lu2022lgdn, dzabraev2021mdmmt, lei2022revealing, li2022lavender, cheng2022vindlu, wang2022language, zhu2020actbert} which leverage the frame-level (or video-level) visual features and fine-grained alignment \cite{zou2022tokenflow, lee2018stacked, messina2021fine,yao2021filip, gou2022leveraging} which utilize the information of patch within each video frames. Recently, several coarse-grained CLIP-based methods \cite{luo2022clip4clip, bain2022clip, gorti2022x} use frame aggregation strategies to convert frame features to video features and perform coarse-grained alignment between the video and query features. Follow-up works \cite{ma2022x, liu2022ts2} investigate various similarity calculation schemes for better cross-modal alignment. TS2-Net \cite{liu2022ts2} designs a cross-modal alignment model between the frame feature and sentence feature of the text query. X-CLIP \cite{ma2022x} utilizes a cross-grained alignment between coarse-grained video features and text features, including video-sentence, video-word, frame-sentence, and frame-word contrast. However, the coarse-grained alignment fails to capture detailed correspondence to due the limited information within high-level features. To this end, TokenFlow \cite{zou2022tokenflow} proposes a fine-grained alignment function for token-wise similarity calculation. The fine-grained alignment does capture more subtle correspondence between text and video, but it could overlook the high-level information like scene and action. In this work, we aim to combine the advantage of both coarse-grained and fine-grained alignment to better capture the correspondence between text query and video candidates.  

\noindent \textbf{Normalization for Video-text Retrieval.}
To retrieve the most relevant video candidate given a text query, common video-text retrieval models compute a similarity matrix between video and text input and retrieve the candidate with the highest similarity. Several previous works \cite{cheng2021improving, bogolin2022cross, park2022normalized} focus on the normalization of this similarity matrix to improve the performance. CAMoE \cite{cheng2021improving} introduces a Dual Softmax Loss (DSL) as a reviser to correct the similarity matrix and achieve the dual optimal match. Later, NCL \cite{park2022normalized} reveals that cross-modal contrastive learning suffers from incorrect normalization of the sum retrieval probabilities of each text or video instance and proposes Normalized Contrastive Learning that computes the instance-wise biases that properly normalize the sum retrieval probabilities. Empirically, we find that the logits from the multi-level similarity matrix are imbalanced and make some videos and texts over- or under-representative. To mitigate the issue, we propose to balance the multi-level alignments by separately normalizing each similarity matrix and aggregating the normalized matrix for better retrieval prediction.

\begin{figure*}[t]
\begin{center}
   \includegraphics[width=1\linewidth]{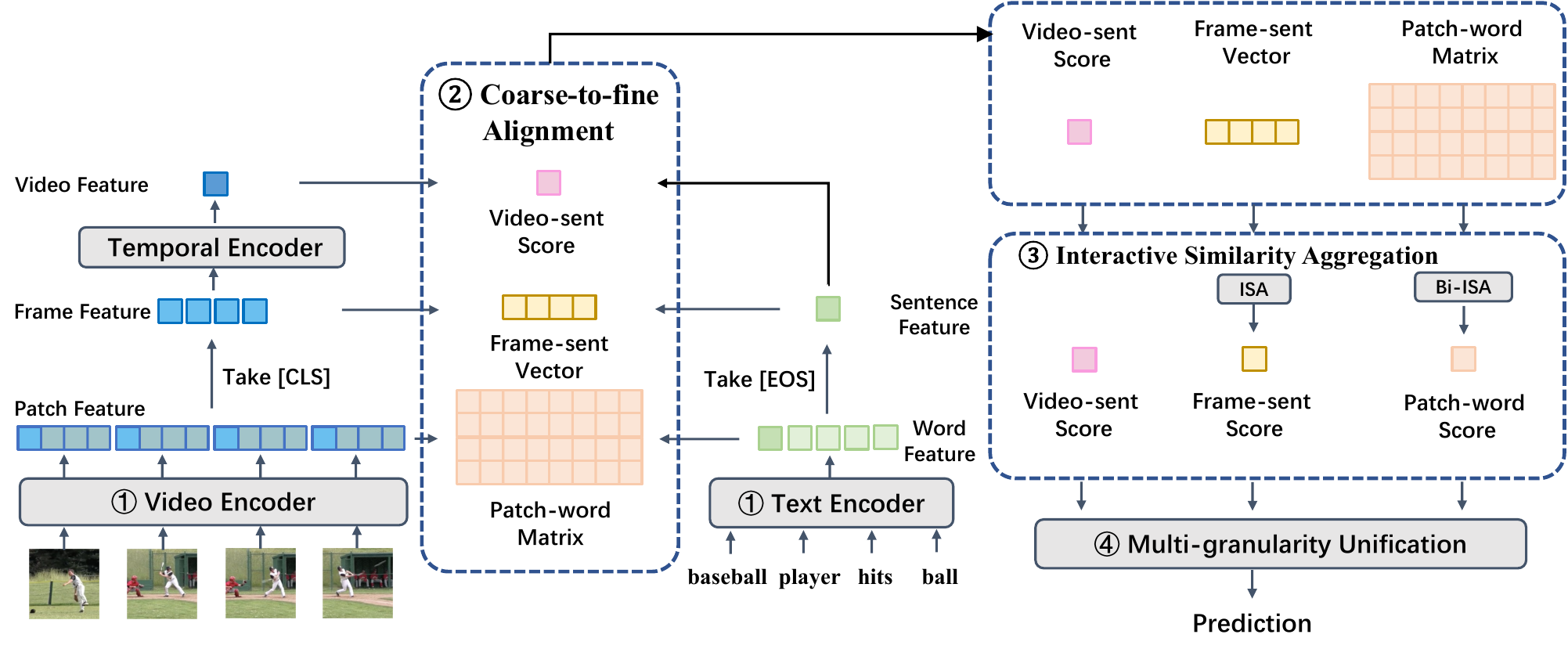}
\end{center}
   \caption{The overview of \Modelname. It consists of four components: (1) text and video encoders, (2) the coarse-to-fine alignment module which consists of video-sentence alignment (top middle), frame-sentence alignment (center), and patch-word alignment (bottom middle), (3) the interactive similarity aggregation module (including ISA for frame-sentence vector and Bidirectional-ISA (Bi-ISA) for patch-word matrix), and (4) the multi-granularity unification module to aggregate the similarity score from different granularity. (For simplicity, ``video-sent'' stands for video-sentence, and frame-sent stands for frame-sentence). 
   \label{fig:architecture}
   }
\end{figure*}

\section{Methodology}
In this selection, we present our proposed \Modelname{} model. As shown in \Cref{fig:architecture}, \Modelname{} consists of four components: (1) text and video encoders, (2) coarse-to-fine alignment module, (3) Interactive Similarity Aggregation module, (4) and multi-granularity unification module with the Sinkhorn-Knopp algorithm. First, the video and text encoders extract multi-grained visual and textual features. Afterward, multi-grained features are fed into a coarse-to-fine alignment module that calculates the different levels of cross-modal similarity. Then, the Interactive Similarity Aggregation module fuses the similarity vector (or matrix, depending on the input type) and obtains the similarity score for each granularity. Finally, the multi-granularity unification module aggregates the similarity scores from all granularity and obtains the final unified similarity score for retrieval prediction. Below, we discuss each of these components in more detail. 

\subsection{Feature Extraction}
\noindent \textbf{Text Encoder.}
Given a text query ${T}$ (we prepend a [EOS] token to $T$), we leverage the CLIP~\cite{radford2021learning} text encoder $\mathcal{F}_t$ to output the word feature $w$, where $w = \mathcal{F}_t \left( T \right)  \in \mathbb{R}^{L_t \times C} $, $L_t$ denotes the length of word sequences, and $C$ denotes the dimension of the word feature. Then we take the representation of the [EOS] token as the sentence feature $s \in \mathbb{R}^{C}$. 

\noindent \textbf{Video Encoder.}
Following \cite{luo2022clip4clip, liu2022ts2}, we utilize the pretrained CLIP visual encoder~\cite{dosovitskiy2020image} ($\mathcal{F}_v$) to extract the visual features of each video. A video with $N$ frames can be denoted as $V = \left[ F_1, F_2, ..., F_N \right]$. Given the $n$-th frame of the video $F_n$,  we divide it into disjoint patches, prepend a [CLS] token to it, and use the vision encoder $\mathcal{F}_v$ to obtain the patch representation ${p}_n$, where ${p}_n = \mathcal{F}_v \left( F_n \right)  \in \mathbb{R}^{M  \times C} $, $M$ denotes the number of the visual patches within a video frame. Note that both textual and visual tokens are embedded in the same dimension $C$. Then we take the [CLS] representation from each frame and combine them together to get the frame representation ${f} = [f_1; f_2; ...; f_N], f \in \mathbb{R}^{N \times C}$. Since we feed frames to ViT separately for better efficiency, the vision encoder cannot learn the temporal information across different frames. To enable the model to learn temporal information with minimal cost, inspired by~\cite{liu2022ts2}, we adopt a token shift module, which shifts the whole spatial token features back and forth across adjacent frames, in the last two transformer blocks of the ViT model.

\subsection{Coarse-to-fine Alignment} \label{ssec: align}
Our proposed coarse-to-fine alignment module calculates the cross-modal similarity from multi-grained visual and textual inputs to address the weakness of only considering either coarse alignment or fine-grained alignment (as shown in \Cref{fig:intro}). First, we adopt a video-sentence alignment to obtain the similarity score of video and sentence features. Then, we leverage a frame-sentence alignment to capture the similarity between each frame and text query and obtain a frame-sentence vector. Lastly, we apply the most fine-grained patch-word alignment to model the similarity between each patch and word representation and obtain a patch-word matrix. Below, we describe each of these alignments in more detail. 

\noindent \textbf{Video-sentence Alignment.} 
To obtain the video representation, following \cite{luo2022clip4clip}, we leverage a temporal encoder to aggregate the frame features $f$ and obtain the video feature $v$. We then compute the cosine similarity between $v \in \mathbb{R}^{C}$ and the sentence feature $s \in \mathbb{R}^{C}$ and get the video-sentence similarity score ${s_{\textsc{vs}}}$. 

\noindent \textbf{Frame-sentence Alignment.}
To obtain the cross-modal similarity between frames and the sentence, we separately compute the cosine similarity between each row in the frame feature $f \in \mathbb{R}^{N \times C}$ and the sentence feature $s \in \mathbb{R}^{C}$ and get the frame-sentence similarity vector ${\textbf{c}_{\textsc{fs}}} \in \mathbb{R}^{N}$.

\noindent \textbf{Patch-word Alignment.}
We then consider the fine-grained alignment between text and video. Since the high-level features fail to convey detailed cross-modal information, utilizing the low-level features helps the model to capture subtle correspondence between text and video. To match the feature granularity, we utilize patch and word features for fine-grained alignment. However, due to the high redundancy of patch features, it is ineffective to align all patches to words. Inspired by \cite{liu2022ts2}, we adopt an MLP-based patch selection module $\mathcal{H}$ to select the top-$K$ salient patches from each frame according to the frame and video feature that correspond to the patch. The selected patch feature can be denoted as $\hat{p} = \mathcal{H}\left(p,f,v\right) \in \mathbb{R}^{L_v \times C}$, where $L_v = N*K$. Afterward, by computing the cosine similarities between all combinations in rows in $\hat{p}$ and rows in the word feature $w \in \mathbb{R}^{L_t \times C}$, we obtain the patch-word similarity matrix ${\textbf{C}_{\textsc{pw}}} \in \mathbb{R}^{L_v \times L_t}$.  

Overall, the coarse-to-fine alignment module allows our model to capture cross-modal similarity from the different granularity of features. In \Cref{table:abla1}, we demonstrate the effectiveness of each level alignment quantitatively.

\subsection{Interactive Similarity Aggregation} \label{ssec: isa}
Next, we describe how we aggregate the cross-modal similarity vector and matrix for different alignments. Due to the high redundancy of video features, irrelevant information within the similarity vector could impede the model from learning precise cross-modal correspondence. Existing methods \cite{liu2022ts2} propose a softmax-based weighted combination of the similarity vector to reduce the impact of irrelevant information. However, the softmax weights fail to capture the information between input features. For instance, while aggregating the frame-sentence alignment, softmax weights ignore the temporal information across frames. As a result, the weighted combination only focuses on the cross-modal relevance between text query and video frames and ignores the interaction between different frames. To this end, we propose a simple, yet effective interactive similarity aggregation module (ISA). 

\begin{figure}[t]
\begin{center}
   \includegraphics[width=1\linewidth]{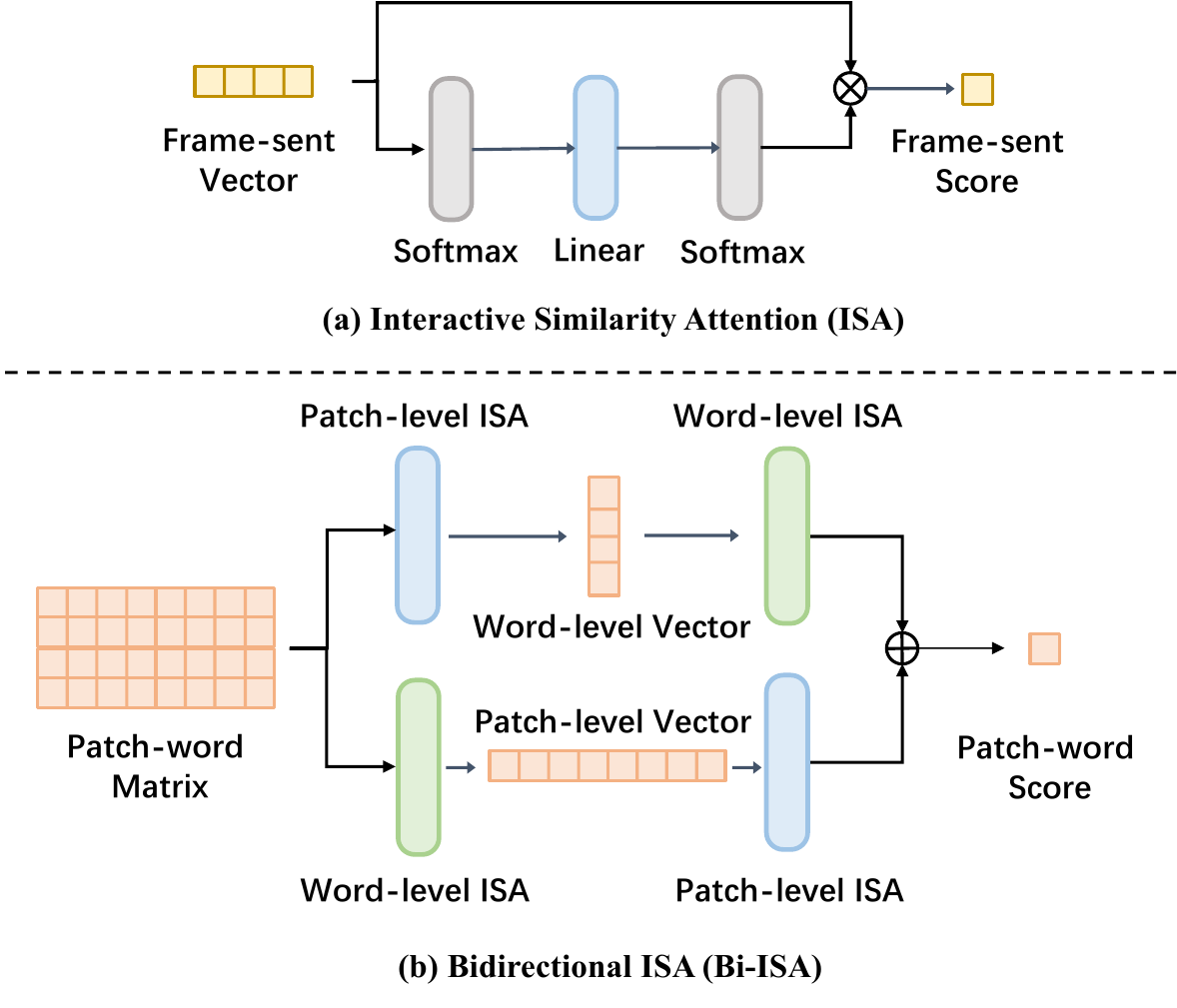}
\end{center}
   \caption{Interactive Similarity Aggregation module (ISA). (a) We directly leverage the ISA module to aggregate the frame-sentence vector to obtain the frame-sentence score. (b) Further, we extend the ISA module to bidirectional ISA (Bi-ISA) to aggregate the patch-word matrix. (For simplicity, frame-sent stands for frame-sentence).}
\label{fig:isa}
\end{figure}

The idea of the ISA module is to jointly consider the cross-modal relevance and the interaction between different features while calculating the weights of different similarities. Inspired by \cite{liu2022ts2, ma2022x}, we first compute the cross-modal relevance by applying a softmax layer on the similarity vector. We then adopt a linear layer $\mathcal{T}_i$ on the cross-modal relevance to encourage the interaction between different features. Finally, we apply another softmax layer to obtain the final weight of the similarity vector. ISA module enables the model to better focus on the related features while aggregating the similarity vector. Specifically, given frame-sentence similarity vector ${\textbf{c}_{\textsc{fs}}} \in \mathbb{R}^{N}$, the ISA module can be formulated as: 

\begin{equation}
{s_{\textsc{fs}}} = \text{Softmax} \left( \mathcal{T}_i \left( \text{Softmax} \left( {\textbf{c}_{\textsc{fs}}} \right)\right) \right) {\textbf{c}_{\textsc{fs}}} , 
\end{equation}
where ${s_{\textsc{fs}}}$ denotes the frame-sentence similarity score between the text query and video candidate. To sum up, the ISA module is capable of aggregating the similarity vector to obtain a similarity score by jointly considering cross-modal relevance and feature interaction. Regardless of the feature dimension, the ISA module is flexible enough to deal with similarity vectors with different feature granularity. Thus, we extend the ISA module to aggregate the patch-word similarity matrix.

A direct idea is to flatten the patch-word matrix to a large vector and apply the ISA module to obtain the similarity score. However, due to the modality gap, it is difficult to model the feature interaction across video and text for the ISA module. The alternative idea is to split each row or column of the similarity matrix into a similarity vector and leverage the ISA module on each vector. Afterward, we aggregate the similarity score from each vector to another similarity vector and apply another ISA to obtain the patch-word score. In that way, we can separately model the feature interaction between patches and words to provide better similarity aggregation. We consider two ways of aggregation: patch-then-word (see the top of \Cref{fig:isa}(b)) and word-then-patch (see the bottom of \Cref{fig:isa}(b)). Empirically we find that jointly considering these two directions provides better aggregation for the patch-word matrix. To this end, we combine the strength of both directions and name the module as a bi-directional ISA module (Bi-ISA, \Cref{fig:isa}(b)). To describe the module formally, we first adopt a patch-level ISA module $\mathcal{A}_p$ on ${\textbf{C}_{\textsc{pw}}}$ to obtain a word-level similarity vector. We then adopt a word-level ISA module $\mathcal{A}_w$ to aggregate the word-level similarity vector to the patch-then-word score. Similarly, we can obtain the word-then-patch score by leveraging a word-level ISA module and a patch-level ISA in a reverse way. The whole process of Bi-ISA can be formulated as: 

\begin{equation}
{s_{\textsc{pw}}}=  \mathcal{A}_p \left( \mathcal{A}_w \left( {\textbf{C}_{\textsc{pw}}} \right)\right) + \mathcal{A}_w \left( \mathcal{A}_p \left({\textbf{C}_{\textsc{pw}}} \right)\right),
\label{eq:Bi-ISA}
\end{equation}

where ${s_{\textsc{pw}}}$ denotes the patch-word similarity score. Conceptually, our ISA module jointly considers the cross-modal relevance and the interaction between different features while aggregating the similarity vector (matrix). In \Cref{table:abla3-1} and \Cref{table:abla3-2}, we validate the effectiveness of our ISA and Bi-ISA module compared to different aggregation mechanisms. In the next section, we further aggregate the different levels of similarity score to one final score for retrieval.

\subsection{Unifying Coarse and Fine-grained Alignments} \label{ssec: unify}

For simplicity, we explained our approach above with only one video-query pair; but when we perform retrieval on $G$ videos and $H$ queries, we compute the similarity scores over all possible combinations of the videos and queries, and denote the score combinations for each alignment level (video-sentence, frame-sentence, patch-word) as $\textbf{S}_{\textsc{vs}}, \textbf{S}_{\textsc{fs}}, \textbf{S}_{\textsc{pw}} \in \mathbb{R}^{G \times H}$, where $\textbf{S}^{ij}$ is the score for the $i^{th}$ video and $j^{th}$ query. For the last step of obtaining the final cross-modal similarity for retrieval, the common method \cite{ma2022x} is usually to directly compute the average over the different levels of similarities. 

However, we find that scores across different videos are highly imbalanced in the similarity matrices of each level, and we empirically find that correcting the issue before summing the similarities leads to a better result in multi-level alignment methods \cite{ma2022x, liu2022ts2}. The imbalance issue is similar to the findings of Park \etal~\cite{park2022normalized}. Specifically, sometimes the summation of retrieval similarities between one specific video and all texts (we called this term \textit{marginal similarity} in the following) might be much higher than that of the other videos, meaning that this video is over-represented and will lower the probability of the other video being selected. To address this, we re-scale the similarity matrix to normalize the marginal similarity of every video to be a similar value. One approach is to apply dual softmax operation \cite{cheng2021improving} on the similarity matrix, but this is not realistic in the testing phase since it requires obtaining all the testing videos and queries at hand.

A more realistic setting is where we can access all the testing videos, but only have one test query at a time for text-to-video retrieval(reversely, for video-to-text retrieval, we search for the best candidate out of all testing queries for one test video). Since there is only one data point on the query side, double softmax is not useful in this case. To address this, we use the training query set as the approximation for the test set and follow Park \etal~\cite{park2022normalized} to leverage the Sinkhorn-Knopp algorithm \cite{cuturi2013sinkhorn} to correct the imbalance. We describe our algorithm for text-to-video retrieval for simplicity as we can just swap videos and texts in the algorithm for video-to-text retrieval.
Specifically, assuming we have $G$ test videos and $J$ train queries, for each level of similarity score, the algorithm computes the test video bias $\alpha \in \mathbb{R}^G$ and training text bias $\beta \in \mathbb{R}^J$ in an alternating and iterative manner. Please refer to the supplements for the details of the algorithm. Adding these biases to the similarity matrix can normalize it to have similar marginal similarities for videos and for texts, respectively. However, since we only introduce the training query set to approximate the test set distribution and help to compute the test video bias $\alpha$, we discard the training query bias $\beta$ after. We then add $\alpha$ to the testing similarity matrix $\textbf{S} \in \mathbb{R}^{G \times H}$ that is generated by $G$ test videos and $H$ test queries, that is,

\begin{equation}\label{eq: sk}
\text{SK}(\textbf{S})^{ij} = \textbf{S}^{ij} + \alpha^i
\end{equation}

We then obtain a normalized similarity matrix for testing videos. We apply the algorithm separately on the similarity matrix of different alignments before summing them together, and we empirically find out this is better than doing summation first and then normalization. Finally, the final retrieval score $\textbf{R}$ can be written as: 

\begin{equation} \label{eq: final score}
\textbf{R}=\text{SK}(\textbf{S}_{\textsc{VQ}}) + \text{SK}(\textbf{S}_{\textsc{SQ}}) +\text{SK}(\textbf{S}_{\textsc{PQ}})
\end{equation}

Note that we only apply \Cref{eq: sk,eq: final score} in the inference phase. Similarly, for video-to-text retrieval, we normalize the similarity matrix by adding the test text bias. In \Cref{table:abla4}, we validate the effectiveness of applying the above normalization. We also provide a visualization of the reduction of over-/under-representation by applying this technique in the supplementary material.

\subsection{Training and Inference} \label{ssec: training and inference}

During training, we randomly sample $B$ video-query pairs, compute the scores over all possible combinations between the videos and the queries without normalization, and denote the similarity combination as $\textbf{R} \in \mathbb{R}^{B \times B}$, where $\textbf{R}^{ij}$ is the score for the $i^{th}$ video and $j^{th}$ query. We utilize the cross-modal contrastive objective~\cite{radford2021learning} to maximize the scores of the positive pairs (the diagonal in $\textbf{R}$) and minimize the scores of negative pairs:

\begin{equation}
\mathcal{L}_{v 2 t}=-\frac{1}{B} \sum_i^B \log \frac{\exp \left(\textbf{R}^{ii}\right)}{\sum_{j=1}^B \exp \left(\textbf{R}^{ij}\right),}
\end{equation}

\begin{equation}
\mathcal{L}_{t 2 v}=-\frac{1}{B} \sum_i^B \log \frac{\exp \left(\textbf{R}^{ii}\right)}{\sum_{j=1}^B \exp \left(\textbf{R}^{ji}\right),}
\end{equation}

\begin{equation}
\mathcal{L}=\mathcal{L}_{v 2 t}+\mathcal{L}_{t 2 v},
\end{equation}

During inference, to perform video-text retrieval, we will compute the similarities between all videos and the query, normalize the similarities by the method introduced in \Cref{ssec: unify}, and retrieve the video with the highest similarity. We conduct the procedure similarly but in the other direction for video-to-text retrieval.

\section{Experimental Setup}

\subsection{Datasets}
We evaluate \Modelname{} on five popular video-text retrieval datasets: MSR-VTT \cite{xu2016msr}, MSVD \cite{chen2011collecting}, ActivityNet \cite{krishna2017dense} and DiDeMo \cite{anne2017localizing}. 

\textbf{MSR-VTT} \cite{xu2016msr} contains $10,000$ videos, each annotated with $20$ text captions. The video length is ranged from $10$ to $32$ seconds. Following \cite{liu2019use, gabeur2020multi}, we train \Modelname{} on $9,000$ videos and report the results on $1,000$ selected video-text pairs (the 1kA test set).

\textbf{Activity-Net} \cite{krishna2017dense} consists of $20,000$ YouTube videos with $100,000$ captions. The average video length is $180$ seconds. We follow \cite{luo2022clip4clip} to concatenate the multiple text descriptions of a video into one paragraph and perform paragraph-to-video retrieval on ActivityNet. We train our model on $10,000$ videos and use the 'val1' split for evaluation which contains $5,000$ videos. 

\textbf{DiDeMo} \cite{anne2017localizing} is comprised of $10,000$ videos and $40,000$ captions. The average video length is 30 seconds. Similar to ActivityNet, we evaluate paragraph-to-video retrieval on DiDeMo. There are $8,395$ videos in the training set, $1,065$ videos in the validation set, and $1,004$ videos in the test set. We report the results on the test set for evaluation.

\textbf{MSVD} \cite{chen2011collecting} contains $1,970$ videos, each with a length that ranges from $1$ to $62$ seconds. Each video contains approximately $40$ captions. Following \cite{luo2022clip4clip}, we split the train, validation, and test set with $1,200$, $100$, and $670$ videos. We follow the common multiple caption evaluation \cite{liu2019use, luo2022clip4clip} setting in which each video in the test set is associated with multiple text captions. 

\textbf{VATEX} \cite{wang2019vatex} consists of $34,991$ video clips with multiple captions per video. We follow HGR's \cite{chen2020fine} split protocol. There are $25,991$ videos in the training set, $1,500$ videos in the validation set and $1,500$ videos for evaluation.

\begin{table*}[]
\centering
\setlength{\tabcolsep}{3pt}

\begin{tabular}{cll|ccc|ccc|ccc|ccc|ccc}
\multicolumn{3}{c|}{\multirow{2}{*}{Method}}        & \multicolumn{3}{c|}{MSR-VTT}   & \multicolumn{3}{c|}{Activity-Net} & \multicolumn{3}{c|}{DiDeMo}     & \multicolumn{3}{c|}{MSVD}  & \multicolumn{3}{c}{VATEX}     \\ \cline{4-18} 

\multicolumn{3}{c|}{}          & R@1      & R@5      & MnR$\downarrow$      & R@1        & R@5       & MnR$\downarrow$      & R@1      & R@5      & MnR$\downarrow$      & R@1      & R@5      & MnR$\downarrow$  & R@1      & R@5      & MnR$\downarrow$   \\ \hline

\multicolumn{18}{c}{Non-CLIP Methods} \\ \hline

\multicolumn{3}{c|}{CE \cite{liu2019use}}                  & 20.9     & 48.8     & 28.2     & 18.2       & 47.7      & 23.1     & 16.1     & 41.1     & 43.7      & 19.8     & 49.0     & -   & -  & - & -  \\
\multicolumn{3}{c|}{MMT \cite{gabeur2020multi}}            & 26.6     & 57.1     & 24.0     & 26.6       & 57.1      & 16.0     & -        & -        & -         & -        & -        & -    & -  & - & -   \\
\multicolumn{3}{c|}{Support Set \cite{patrick2020support}} & 30.1     & 58.5     & -        & 29.2       & 61.6      & -        & -        & -        & -         & 28.4     & 60.0     & -   & 45.9 & 82.4 & -    \\
\multicolumn{3}{c|}{Frozen \cite{bain2021frozen}}          & 31.0     & 59.5     & -        & -          & -         & -        & 34.6     & 65.0     & -         & 33.7     & 64.7     & -   & -  & - & -  \\

\multicolumn{3}{c|}{All-in-one \cite{wang2022all}}          & 37.9    & 68.1     & -        &  22.4       & 53.7       & -        & 32.7    & 61.4     & -         & -    & -     & -  & -  & - & -   \\

\multicolumn{3}{c|}{\textcolor{lightgray}{\textit{Singularity \cite{lei2022revealing}}}}          & \textcolor{lightgray}{\textit{42.7}}     & \textcolor{lightgray}{\textit{69.5}}     & \textcolor{lightgray}{\textit{-}}       & \textcolor{lightgray}{\textit{48.9}}          & \textcolor{lightgray}{\textit{77.0}}        & \textcolor{lightgray}{\textit{-}}        & \textcolor{lightgray}{\textit{53.1}}     & \textcolor{lightgray}{\textit{79.9}}     & \textcolor{lightgray}{\textit{-}}         & \textcolor{lightgray}{\textit{-}}     & \textcolor{lightgray}{\textit{-}}   & \textcolor{lightgray}{\textit{-}}   & \textcolor{lightgray}{\textit{-}}  & \textcolor{lightgray}{\textit{-}} & \textcolor{lightgray}{\textit{-}}  \\

\multicolumn{3}{c|}{\textcolor{lightgray}{\textit{VindLU \cite{cheng2022vindlu}}}}  & \textcolor{lightgray}{\textit{45.3}}     & \textcolor{lightgray}{\textit{69.9}}     & \textcolor{lightgray}{\textit{-}}        & \textcolor{lightgray}{\textit{54.4}}     & \textcolor{lightgray}{\textit{80.7} }    & \textcolor{lightgray}{\textit{-}} & \textcolor{lightgray}{\textit{59.2}}          & \textcolor{lightgray}{\textit{84.1}}        & \textcolor{lightgray}{\textit{-}}        &    \textcolor{lightgray}{\textit{-}}     & \textcolor{lightgray}{\textit{-}}   & \textcolor{lightgray}{\textit{-}}     & \textcolor{lightgray}{\textit{-}}  & \textcolor{lightgray}{\textit{-}}  & \textcolor{lightgray}{\textit{-}}  \\\hline

\multicolumn{18}{c}{CLIP-based Methods} \\ \hline

\multicolumn{3}{c|}{CLIP4Clip \cite{luo2022clip4clip}}     & 44.5     & 71.4     & 15.3     & 40.5       & 73.4      & 10.0     & 43.4     & 70.2     & 17.5      & 46.2     & 76.1     & 10.0   & 55.9 & 89.2 & 3.9  \\
\multicolumn{3}{c|}{CAMoE \cite{cheng2021improving}}     & 44.6     & 72.6      & 13.3     & -       &-      & -    & -   & -     & -     & 46.9     & 76.1      & 9.8  & -        & -        & -  \\
\multicolumn{3}{c|}{X-Pool \cite{gorti2022x}}              & 46.9     & 72.8     & 14.3     & -          & -         & -        & -        & -        & -         & 47.2     & 77.4     & \textbf{9.3}  & -        & -        & -\\
\multicolumn{3}{c|}{TS2-Net \cite{liu2022ts2}}             & 47.0     & \textbf{74.5} & 13.0     & 41.0       & 73.6      & 8.4      & 41.8     & 71.6     & 14.8      & -        & -        & -    & 59.1 & 90.0  & 3.5    \\
\multicolumn{3}{c|}{X-CLIP \cite{ma2022x}}                 & 46.1     & 73.0     & 13.2     & 44.3       & 74.1      & 7.9      & 45.2     & 74.0     & 14.6      & 47.1     & \textbf{77.8} & 9.5    & -        & -        & -  \\ \hline
\multicolumn{3}{c|}{\Modelname{}}                         & \textbf{49.4} & 72.1     & \textbf{12.9} & \textbf{45.7}   & \textbf{76.0}  & \textbf{6.6}  & \textbf{46.5} & \textbf{74.8} & \textbf{13.4} & \textbf{47.4} & 77.6     & 9.6   & \textbf{61.1} & \textbf{90.5}  & \textbf{3.4} \\

\end{tabular}
\caption{Comparison to the state-of-the-art text-to-video retrieval methods on MSR-VTT, AcitivityNet, DiDeMo, MSVD, VATEX. The top section shows the results of non-CLIP methods and the middle section shows the results of CLIP-based methods. Our results indicate that \Modelname{} achieves better or comparable results on all five datasets compared to the current state-of-the-art methods. For a fair comparison, we de-emphasize Singularity~\cite{lei2022revealing} and VindLU~\cite{cheng2022vindlu} (by using gray color and italic font) since they are pretrained on large-scale video datasets and use time-consuming two-stage re-ranking strategy (two-step retrieval, the first step retrieves top-$K$ candidates from all candidates and the second step retrieves the best candidate from the top-$K$ candidates).}
\label{table:sota}
\end{table*}

\begin{table}[]
\centering
\setlength{\tabcolsep}{2.3pt}
\begin{tabular}{cl|cc|cc|cc}
\multicolumn{2}{c|}{\multirow{2}{*}{Method}}        & \multicolumn{2}{c|}{MSR-VTT}   & \multicolumn{2}{c|}{Activity-Net} & \multicolumn{2}{c}{DiDeMo}    \\ \cline{3-8} 

\multicolumn{2}{c|}{} & R@1  & R@5 & R@1  & R@5 & R@1  & R@5 \\ \hline
\multicolumn{2}{c|}{CE \cite{liu2019use}}               & 20.6 & 50.3 & 17.7 & 46.6 & 15.6 & 40.9\\
\multicolumn{2}{c|}{MMT \cite{gabeur2020multi}}                   & 27.0 & 57.5 & 28.9 & 61.1 & -   & -\\
\multicolumn{2}{c|}{Support set \cite{patrick2020support}}              & 26.6 & 55.1 &28.7 & 60.8 & -   & -\\
\multicolumn{2}{c|}{HiT \cite{liu2021hit}}                    & 32.1 & 62.7 & -    & - & -   & - \\ 
\multicolumn{2}{c|}{TT-CE \cite{croitoru2021teachtext}}                    & 32.1 & 62.7 & 23.0 & 56.1 & 21.1 & 47.3  \\ \hline
\multicolumn{2}{c|}{CLIP4Clip \cite{luo2022clip4clip}}   & 42.7 & 70.9 & 42.5 & 74.1   & 42.5 & 70.6 \\
\multicolumn{2}{c|}{TS2-Net \cite{liu2022ts2}}           & 45.3 & 74.1 & -    & -  & -    & - \\
\multicolumn{2}{c|}{X-CLIP \cite{ma2022x}}             & 46.8 & 73.3 & 43.9 & 73.9 &43.1 & \textbf{72.2}\\
\hline
\multicolumn{2}{c|}{\Modelname{}}           & \textbf{47.1} & \textbf{74.3}  & \textbf{46.3} & \textbf{76.5} &\textbf{46.0} & 71.9 
\end{tabular}
\caption{Comparison to the state-of-the-art video-to-text retrieval methods on MSR-VTT, AcitivityNet, DiDeMo datasets. The top section shows the results of non-CLIP methods and the middle section shows the results of CLIP-based methods. Our results indicate that \Modelname{} achieves better or comparable results on all datasets compared to the current state-of-the-art methods.}
\label{table:v2t}
\end{table}

\subsection{Evaluation Metrics}
Following \cite{luo2022clip4clip}, we use standard video-text retrieval metrics, including R@1, R@5, and Mean Rank (MnR) to validate the effectiveness of our \Modelname{} model.  
We report results with more metrics (including R@10, Median Recall) in supplements. 

\subsection{Implementation Details}
Following \cite{luo2022clip4clip}, we leverage the CLIP \cite{radford2021learning} pre-trained weights to initialize our \Modelname{} model. For the visual encoder, we use CLIP's ViT-B/32 weights. The dimension $C$ of visual and textual representations is set to $512$. We choose the $K=4$ most salient patches for each frame in our patch selection module on all datasets. For MSR-VTT, MSVD, and VATEX, we follow the previous work \cite{luo2022clip4clip} to sample $N =12$ frames per video and set the max length of text query as $32$. For the paragraph-to-video dataset ActivityNet and DiDeMo, we sample $64$ frames per video and set the max length of text query as $64$. We adopt the Adam optimizer \cite{kingma2014adam} with a cosine warm-up strategy \cite{loshchilov2016sgdr}. We set the learning rate of the visual and text encoders as $1\mathrm{e}{-7}$ and other modules as $1\mathrm{e}{-4}$. In the Sinkhorn-Knopp algorithm, we set the number of iterations as $4$ across all datasets. We set a batch size to 128 for MSR-VTT, MSVD, and VATEX and 64 for ActivityNet and Didemo following \cite{ma2022x}. We train \Modelname{} for $8,5,20,20,20$ epochs on MSR-VTT, MSVD, ActivityNet, DiDeMo, and VATEX, respectively. We conduct the ablation study on the most popular MSR-VTT dataset to analyze the effect of different designs of our model.

\section{Experimental Results}

In this section, we compare \Modelname{} with several recent methods on the five video-text retrieval datasets and conduct comprehensive ablation studies to verify our design choices. We also provide a qualitative analysis to show the effectiveness of our model designs.  Moreover, we display quantitative results with full metrics (R@1,5,10, MdR, MnR) for each dataset, more experiments about applying \Modelname{} to more advanced backbone model \cite{xue2022clip}, and quantitative analysis on computational cost and training strategy in the supplements.

\subsection{Comparison to State-of-the-art Approaches}
\label{ssec:comparison to sota}

In \Cref{table:sota}, we compare \Modelname{} with existing methods that are either with (in the middle section of the table) or without (in the upper section of the table) CLIP on text-to-video retrieval. We also compare \Modelname{} with existing methods on video-to-text retrieval in \Cref{table:v2t}. We observe that \Modelname{} achieves better performance than existing methods on most of the metrics on both text-to-video and video-to-text retrieval settings. 

On MSR-VTT, compared to the recent multi-level alignment method X-CLIP \cite{ma2022x}, \Modelname{} gives a significant $3.3\%$ improvement on text-to-video R@1 metric and comparable video-to-text retrieval results despite X-CLIP leveraging more levels of coarse alignments (video-sentence, video-word, frame-sentence, and frame-word) than us. This result verifies our motivation that building a coarse-to-fine alignment is useful for video-text retrieval. We also achieve $2.0\%$ improvement on text-to-video R@1 on VATEX dataset.

Similarly, on ActivityNet and DiDeMo datasets, we notice that \Modelname{} is capable of handling longer text queries and achieves observe $5.2\%$ and $4.7\%$ improvement on paragraph-to-video retrieval compared to CLIP4Clip \cite{luo2022clip4clip} which only utilizes video-sentence alignment. Furthermore, we observe $1.4\%$ and $1.3\%$ gain on ActivityNet and DiDeMo compared to X-CLIP \cite{ma2022x} under paragraph-to-video retrieval and $2.4\%$ and $2.9\%$ gain on video-to-paragraph retrieval. These results show the importance of fine-grained correspondence even on retrieval for long videos. 

Meanwhile, our method achieves comparable results on the MSVD dataset which evaluates a multiple-caption setting, which further verifies the generalizability of our model. However, we observe our normalization before the summation strategy still introduces some performance gain on MSVD even though the multiple-caption setting breaks our assumption that one video has one corresponding query, showing the robustness of our approach.

\subsection{Ablation study} \label{abla}
In this section, we study the different design choices of our \Modelname{} model and verify their effects on the video-text retrieval performance on MSR-VTT under text-to-video retrieval setting. Specifically, we investigate (1) the effect of different alignment schemes, (2) the comparison of our fine-grained alignment design to others, (3) different similarity aggregation methods and (4) the effect of the unification module. 

\begin{table}[]
\centering
\setlength{\tabcolsep}{4pt}
\begin{tabular}{ccc|ccc}

video-sent & frame-sent & patch-word & R@1  & R@5  & MnR$\downarrow$  \\ \hline
  \checkmark         &            &            & 44.5 & 71.1 & 15.2 \\
     \checkmark      &     \checkmark       &            & 47.1 & 73.2 & 14.1 \\
      \checkmark     &   \checkmark         &   \checkmark   & \textbf{48.2} & \textbf{73.3} & \textbf{13.2}
\end{tabular}
\caption{The effect of different level alignments. Video-sent denotes the video-sentence alignment, frame-sent denotes the frame-sentence alignment and patch-word denotes patch-word alignment. The results indicate the effectiveness of each level alignment. 
}
\label{table:abla1}

\end{table}

\begin{table}[]
\centering
\begin{tabular}{cc|ccc}

patch-sent & patch-word & R@1  & R@5  & MnR$\downarrow$ \\ \hline
           &            & 47.1 & 73.2 & 14.1 \\
      \checkmark     &            & 46.5 & \textbf{73.7} & 14.7 \\
           &    \checkmark        & \textbf{48.2} & 73.3 & \textbf{13.2}
           \\
    \checkmark       &    \checkmark        & 47.2 & 71.9 &  13.8
\end{tabular}
\caption{Comparison of our fine-grained alignment design to others. ``Patch-word'' denotes patch-word alignment and ``patch-sentence'' denotes patch-sentence alignment. The last row denotes the ensemble of patch-word and patch-sentence alignment. The results indicate our patch-word alignment is the best design choice for fine-grained alignment on MSR-VTT.}
\label{table:abla2}

\end{table}

\begin{table}[]
\centering
\begin{tabular}{ccc|ccc}
\multicolumn{3}{c|}{Aggregation Method} & R@1 & R@5 & MnR $\downarrow$ \\ \hline
\multicolumn{3}{c|}{Mean Pooling} & 45.6  &  71.2  & 15.2        \\
\multicolumn{3}{c|}{Softmax Weight} & 46.2  &  72.7  & 14.7      \\
\multicolumn{3}{c|}{ISA (ours)}         & \textbf{47.1} & \textbf{73.2} & \textbf{14.1}   
\end{tabular}
\caption{Effect of different similarity aggregation methods for frame-sentence vector. Results show that our ISA module has better performance on all metrics. 
}
\label{table:abla3-1}

\end{table}

\noindent \textbf{The Effect of Different Alignment Schemes.} 
First, we validate the effectiveness of our different levels of alignment. As shown in \Cref{table:abla1}, adding frame-sentence and patch-word alignments improves the base model (that only leverages video-sentence alignment) with a significant margin. Specifically, we observe that adding patch-word alignment
improves the R@1 while keeping a similar R@5, which indicates that the fine-grained alignment helps the model choose the most relevant video (top-$1$) from several similar video candidates (top-$5$) by capturing the subtle differences between these video candidates.
This result justifies our motivation for using fine-grained alignment as complementary to coarse alignment.

\noindent \textbf{Comparison of Our Fine-grained Alignment Design to Others.} 
In \Cref{table:abla2}, we study the different designs of fine-grained alignment in our model. We compare our patch-word similarity with patch-sentence alignment
and the ensemble of these two alignments. Based on the results, we observe that our patch-word alignment is the best design, and meanwhile, adding patch-sentence alignment degrades the performance, possibly caused by the mismatch between patch and sentence representations, where one conveys local information and the other contains global information. We also provide qualitative analysis on different alignment designs in supplements.

\noindent \textbf{Different Similarity Aggregation Methods.} 
To validate the effectiveness of our ISA module for frame-sentence alignment and Bi-ISA module for patch-word alignment in \Cref{ssec: isa}, we compare our modules with several other aggregation methods. In \Cref{table:abla3-1}, we show the effectiveness of our interactive similarity attention (ISA) on the frame-sentence score. Specifically, we compare the vanilla mean pooling strategy and softmax-based weighted combination \cite{liu2022ts2}. Note that we only adopt video-sentence and frame-sentence alignment (remove patch-word alignment and the Bi-ISA module) for the experiments in \Cref{table:abla3-1} to study the effect of ISA independently. Results show that using the ISA module achieves better performance compared to other aggregation methods. We also compare our Bi-ISA with other aggregation methods in \Cref{table:abla3-2}, where we have adopted ISA for the frame-sentence score. We observe that the design of bi-directional aggregation achieves a significant gain in performance. In general, our experiments show that adding one linear layer to learn the temporal information across frames before aggregation is crucial.

\begin{table}[]
\centering
\setlength{\tabcolsep}{4pt}
\begin{tabular}{c|c|ccc}
Aggregation Method & Bidirectional & R@1  & R@5  & MnR$\downarrow$ \\ \hline
Direct-ISA         &               & 46.4 & 72.8 & 14.8 \\
Mean Pooling       &     \checkmark          & 47.1 & 73.0 & 13.9 \\
Softmax Weight     &     \checkmark          & 47.5 & \textbf{73.4} & 13.5 \\
Bi-ISA (ours)      &      \checkmark         & \textbf{48.2} & 73.3 & \textbf{13.2}
\end{tabular}
\caption{Effect of different similarity aggregation methods for patch-word matrix. Results verify the effectiveness of our Bi-ISA module compared to other methods. }
\label{table:abla3-2}
\end{table}

\noindent \textbf{The Effect of Unification Module.}
To verify the importance of the unification module for different levels of similarity, we compare our \Modelname{} model with the variant that removes the  Sinkhorn-Knopp normalization (abbreviated as SK norm) in \Cref{table:abla4}. We notice that adding the SK norm provides better performance on video-text retrieval with a $1.2\%$ gain on both R@1 and R@10. 

\begin{table}[]
\centering

\begin{tabular}{cll|cccc}
\multicolumn{3}{c|}{Setting}   & R@1  & R@5 & R@10 & MnR$\downarrow$  \\ \hline
\multicolumn{3}{c|}{w/o SK norm} & 48.2 & \textbf{73.3} & 82.3 & 13.2 \\
\multicolumn{3}{c|}{w SK norm}   & \textbf{49.4} & 72.1 & \textbf{83.5} & \textbf{12.9}
\end{tabular}
\caption{The effect of SK norm of different level similarity scores.}
\label{table:abla4}
\end{table}

\begin{figure}[t]
\begin{center}
   \includegraphics[width=1\linewidth]{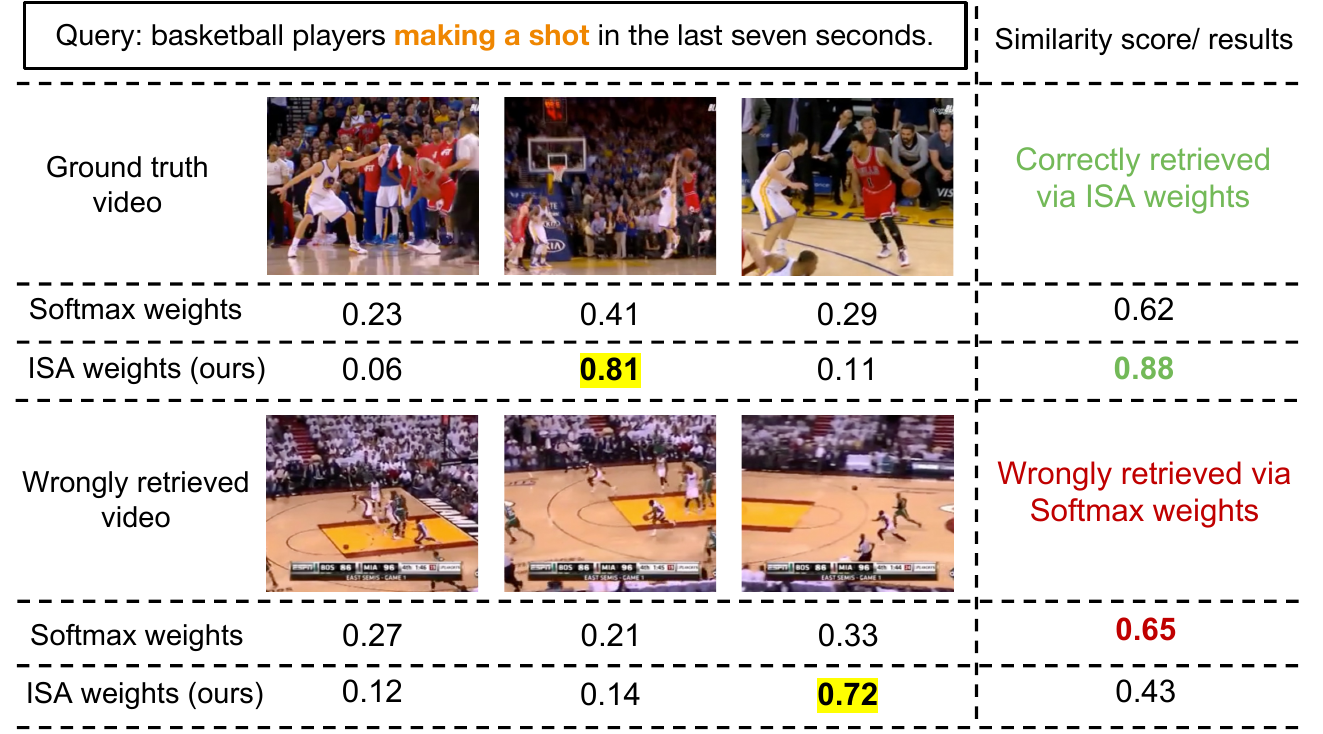}
\end{center}
\setlength \belowcaptionskip{-40pt}
\caption{The visualization of the effectiveness of our ISA module. In the upper part, we show that the ISA improves the softmax weight by highlighting the most relevant frame to the query. As a result, the calculated similarity score of ISA is significantly increased since the irrelevant information is eliminated. In the lower part, we show that the ISA module is also capable of assigning the most related frame (the player is about to get the ball) the highest score for the unmatched video (caption: the player is chasing the ball). However, since it is still not directly related to the text query (compared to the upper video), our model with ISA produces a lower similarity score.}
\vspace{-10pt}
\label{fig:vis}
\end{figure}

\subsection{Qualitative Analysis}
In this section, we report the qualitative analysis of the effectiveness of the ISA module. We show the comparison of the model with and without ISA module in \Cref{fig:vis}. We can see from the upper part of \Cref{fig:vis} that the ISA module improves the softmax weight by highlighting the most relevant frame to the query. As a result, the calculated similarity score of ISA is significantly increased since the irrelevant information is eliminated. In the lower part, by recognizing the most related frame, our model with ISA produces a lower similarity to the unmatched video. For either the ground truth or wrongly retrieved video, we find the softmax weights used in TS2-Net~\cite{liu2022ts2} (without frame-wise interaction) tend to be more uniformly distributed. On the contrary, our ISA module is capable of finding and assigning the highest score to the most relevant frame (the player is making shots) through the interaction between frames, which shows video candidates and results in the correct retrieval. This analysis verifies the effectiveness of our ISA module.

\section{Conclusion}
In this paper, we present \Modelname{}, which jointly considers cross-modal correspondence from different granularity and accomplishes the unification of multi-grained alignment. It achieves state-of-the-art results on multiple video-text retrieval benchmarks. \Modelname{} is a simple but effective model that achieves state-of-the-art results on five diverse video-text retrieval benchmarks. In the future, we plan to extend our method to other video-language tasks such as video question answering and video reasoning.

\section*{Acknowledgment}
We thank the reviewers and Shoubin Yu for their helpful discussions. This work was supported by ARO Award W911NF2110220, ONR Grant N00014-23-1-2356, DARPA KAIROS Grant FA8750-19-2-1004, NSF-AI Engage Institute DRL211263, Sony Faculty Innovation award, and Laboratory for Analytic Sciences via NC State University.

{\small
\bibliographystyle{ieee_fullname}
\bibliography{UCOFIA_arxiv}
}

\appendix
\section*{Appendix}

In this Appendix, we present the following items:

\begin{enumerate}[label=\Alph*.]
    \item Additional Quantitative Results 
    \item Additional Qualitative Results
    \item Method Details
\end{enumerate}

\section{Additional Quantitative Results}

In this section, we report additional quantitative results for our \Modelname{} model. First, we report the results with full video-text retrieval metrics (including video-to-text retrieval) on MSR-VTT, ActivityNet, and DiDeMo. Results indicate our \Modelname{} model achieves better results on both text-to-video and video-to-text retrieval compared to the current state-of-the-art CLIP-based approaches. Meanwhile, we show \Modelname{} is capable of adapting to other advanced backbone models. Then, we compare the performance and computational cost of \Modelname{} with previous work and validate our methods accomplish significant improvement with limited additional computation. Lastly, we ablate the different training settings for encoders and the model design of our bi-directional ISA module (Bi-ISA).

\begin{table*}[]
\centering
\begin{tabular}{cll|ccccc|ccccc}
\multicolumn{3}{c|}{\multirow{2}{*}{Method}} & \multicolumn{5}{c|}{Text $\rightarrow$ Video}        & \multicolumn{5}{c}{Video $\rightarrow$ Text}         \\
\multicolumn{3}{c|}{}                        & R@1  & R@5  & R@10 & MdR$\downarrow$ & MnR$\downarrow$  & R@1  & R@5  & R@10 & MdR$\downarrow$ & MnR$\downarrow$  \\ \hline
\multicolumn{3}{c|}{CE \cite{liu2019use}}                      & 20.9 & 48.8 & 62.4 & 6.0 & 28.2 & 20.6 & 50.3 & 64.0 & 5.3 & 25.1 \\
\multicolumn{3}{c|}{MMT \cite{gabeur2020multi}}                     & 26.6 & 57.1 & 69.6 & 4.0 & 24.0 & 27.0 & 57.5 & 69.7 & 3.7 & 21.3 \\
\multicolumn{3}{c|}{Support set \cite{patrick2020support}}             & 27.4 & 56.3 & 67.7 & 3.0 & -    & 26.6 & 55.1 & 67.5 & 3.0 & -    \\
\multicolumn{3}{c|}{Frozen \cite{bain2021frozen}}                  & 31.0 & 59.5 & 70.5 & 3.0 & -    & -    & -    & -    & -   & -    \\
\multicolumn{3}{c|}{HiT \cite{liu2021hit}}                     & 30.7 & 60.9 & 73.2 & 2.6 & -    & 32.1 & 62.7 & 74.1 & 3.0 & -    \\
\multicolumn{3}{c|}{TT-CE \cite{croitoru2021teachtext}}                   & 29.6 & 61.6 & 74.2 & 3.0 & -    & 32.1 & 62.7 & 75.0 & 3.0 & -    \\ \hline
\multicolumn{3}{c|}{CLIP-straight \cite{portillo2021straightforward}}           & 31.2 & 53.7 & 64.2 & 4.0 & -    & 27.2 & 51.7 & 62.6 & 5.0 & -    \\
\multicolumn{3}{c|}{CLIP4Clip \cite{luo2022clip4clip}}               & 44.5 & 71.4 & 81.6 & \textbf{2.0} & 15.3 & 42.7 & 70.9 & 80.6 & \textbf{2.0} & 11.6 \\
\multicolumn{3}{c|}{CAMoE \cite{cheng2021improving}}                   & 44.6 & 72.6 & 81.8 & \textbf{2.0} & 13.3 & 45.1 & 72.4 & 83.1 & \textbf{2.0} & 10.0 \\
\multicolumn{3}{c|}{X-pool \cite{gorti2022x}}                  & 46.9 & 72.8 & 82.2 & \textbf{2.0} & 14.3 & -    & -    & -    & -   & -    \\
\multicolumn{3}{c|}{X-CLIP \cite{ma2022x}}                  & 46.1 & 73.0 & 83.1 & \textbf{2.0} & 13.2 & 46.8 & 73.3 & \textbf{84.0} & \textbf{2.0} & \textbf{9.1}  \\
\multicolumn{3}{c|}{TS2-Net \cite{liu2022ts2}}                 & 47.0 & \textbf{74.5} & \textbf{83.8} & \textbf{2.0} & 13.0 & 45.3 & 74.1 & 83.7 & \textbf{2.0} & 9.2  \\ \hline
\multicolumn{3}{c|}{\Modelname{}(ViT-32)}                    & \textbf{49.4} & 72.1 & 83.5 & \textbf{2.0} & \textbf{12.9} & \textbf{47.1} & \textbf{74.3} & 83.0 & \textbf{2.0} & 11.4 \\
\multicolumn{3}{c|}{\Modelname{}(ViT-16)}                    & 49.8 & 74.6  & 83.5 & 2.0 & 13.3 &49.1 & 77.0 & 83.8 & 2.0 & 11.2
\end{tabular}
\caption{Comparison to the state-of-the-art video-text retrieval methods on MSR-VTT. The top section shows the results of non-CLIP methods and the middle section shows the results of CLIP-based methods. The bottom section shows the \Modelname{} performance on different size of backbone. For fair comparison, we highlight the best results of each metric using the same backbone model (ViT-$32$).}
\label{table:msrvtt}
\end{table*}

\begin{table*}[]
\centering
\begin{tabular}{cll|ccccc|ccccc}
\multicolumn{3}{c|}{\multirow{2}{*}{Method}} & \multicolumn{5}{c|}{Text $\rightarrow$ Video}          & \multicolumn{5}{c}{Video $\rightarrow$ Text}           \\
\multicolumn{3}{c|}{}       & R@1  & R@5  & R@10 & MdR$\downarrow$ & MnR$\downarrow$ & R@1  & R@5  & R@10 & MdR$\downarrow$ & MnR$\downarrow$ \\ \hline
\multicolumn{3}{c|}{CE \cite{liu2019use}}                      & 18.2 & 47.7 & 91.4 & 6.0             & 23.1            & 17.7 & 46.6 & -    & 6.0             & 24.4            \\
\multicolumn{3}{c|}{MMT \cite{gabeur2020multi}}                     & 28.7 & 61.4 & 94.5 & 3.3             & 16.0            & 28.9 & 61.1 & -    & 4.0             & 17.1            \\
\multicolumn{3}{c|}{Support set \cite{patrick2020support}}             & 29.2 & 61.6 & 94.7 & 3.0             & -               & 28.7 & 60.8 & -    & \textbf{2.0}             & -               \\
\multicolumn{3}{c|}{TT-CE \cite{croitoru2021teachtext}}                   & 23.5 & 57.2 & 96.1 & 4.0             & -               & 23.0 & 56.1 & -    & 4.0             & -               \\ \hline
\multicolumn{3}{c|}{CLIP4Clip \cite{luo2022clip4clip}}               & 40.5 & 72.4 & \textbf{98.2} & \textbf{2.0}             & 7.5             & 42.5 & 74.1 & 85.8 & \textbf{2.0}             & \textbf{6.6}             \\
\multicolumn{3}{c|}{TS2-Net \cite{liu2022ts2}}                 & 41.0 & 73.6 & 84.5 & \textbf{2.0}              & 8.4             & -    & -    & -    & -               & -               \\
\multicolumn{3}{c|}{X-CLIP \cite{ma2022x}}                  & 44.3 & 74.1 & -    & -               & 7.9             & 43.9 & 73.9 & -    & -               & 7.6             \\ \hline
\multicolumn{3}{c|}{\Modelname{}(ours)}                  & \textbf{45.7} & \textbf{76.6} & 86.6 & \textbf{2.0}             & \textbf{6.4}             & \textbf{46.3} & \textbf{76.5} & \textbf{86.3} & \textbf{2.0}              & 6.7            
\end{tabular}
\caption{Video-text retrieval results on ActivityNet.}
\label{table:anet}
\end{table*}

\begin{table*}[]
\centering
\begin{tabular}{cll|ccccc|ccccc}
\multicolumn{3}{c|}{\multirow{2}{*}{Method}} & \multicolumn{5}{c|}{Text $\rightarrow$ Video}       & \multicolumn{5}{c}{Video $\rightarrow$ Text}        \\
\multicolumn{3}{c|}{}                        & R@1  & R@5  & R@10 & MdR$\downarrow$ & MnR$\downarrow$  & R@1  & R@5  & R@10 & MdR$\downarrow$ & MnR$\downarrow$  \\ \hline
\multicolumn{3}{c|}{CE \cite{liu2019use}}                      & 16.1 & 41.1 & -    & 8.3 & 43.7 & 15.6 & 40.9 & -    & 8.2 & 42.4 \\
\multicolumn{3}{c|}{ClipBERT \cite{lei2021less}}                & 20.4 & 48.0 & 60.8 & 6.0 & -    & -    & -    & -    & -   & -    \\
\multicolumn{3}{c|}{TT-CE \cite{croitoru2021teachtext}}                   & 34.6 & 65.0 & 74.7 & 3.0 & -    & 21.1 & 47.3 & 61.1 & 6.3 & -    \\
\multicolumn{3}{c|}{Frozen \cite{bain2021frozen}}                  & 21.6 & 48.6 & 62.9 & 6.0 & -    & -    & -    & -    & -   & -    \\ \hline
\multicolumn{3}{c|}{CLIP4Clip \cite{luo2022clip4clip}}               & 43.4 & 70.2 & 80.6 & \textbf{2.0} & 17.5 & 42.5 & 70.6 & 80.2 & \textbf{2.0} & 11.6 \\
\multicolumn{3}{c|}{TS2-Net \cite{liu2022ts2}}                 & 41.8 & 71.6 & 82.0 & \textbf{2.0} & 14.8 & -    & -    & -    & -   & -    \\
\multicolumn{3}{c|}{X-CLIP \cite{ma2022x}}                  & 45.2 & 74.0 & -    & -   & 14.6 & 43.1 & \textbf{72.2} & -    & -   & \textbf{10.9} \\ \hline
\multicolumn{3}{c|}{\Modelname{}(ours)}                  & \textbf{46.5} & \textbf{74.8} & \textbf{84.4} & \textbf{2.0} & \textbf{13.4} & \textbf{46.0} & 71.9 & \textbf{81.5} & \textbf{2.0} & 12.1
\end{tabular}
\caption{Video-text retrieval results on DiDeMo.}
\label{table:didemo}
\end{table*}

\begin{table}[]
\centering
\begin{tabular}{c|ccc}
Methods  & R@1  & R@5  & R@10 \\ \hline
CLIP-ViP \cite{xue2022clip} &  50.1 &74.8 & 84.6  \\
   \Modelname{} with CLIP-ViP      & \textbf{51.3} & \textbf{75.1} & \textbf{85.2}
\end{tabular}
\caption{Text-to-video retrieval results on MSR-VTT dataset under CLIP-ViP backbone model.}
\label{table:vip}
\end{table}

\begin{table}
\centering
\setlength{\tabcolsep}{4pt}
\begin{tabular}{c|ccc} 
Model     & R@1  & Param (M) $\downarrow$ & Mem (GB) $\downarrow$ \\ \hline
X-CLIP \cite{ma2022x}   & 46.1                & \textbf{164}   & \textbf{12.5}     \\
\Modelname{}(ours)      & \textbf{49.4}                & 166 & 13.9      
\end{tabular}
\caption{The comparison of performance (text-to-video retrieval on MSR-VTT) and computational cost (model parameters and memory) between X-CLIP and \Modelname{}(ours).}
\label{table:computation}
\end{table}

\begin{table}[]
\centering
\setlength{\tabcolsep}{3pt}
\begin{tabular}{cc|ccc}
Patch-then-word & Word-then-patch & R@1  & R@5  & MnR$\downarrow$  \\ \hline
   \checkmark    &                 & 47.8 & 72.8 & 13.4 \\
                &  \checkmark   & 47.7 & 72.9 & 13.5 \\
  \checkmark  &   \checkmark    & \textbf{48.2} & \textbf{73.3} & \textbf{13.2}
\end{tabular}
\caption{The effect of leveraging both aggregation directions (patch-then-word and word-then-patch) for patch-word matrix aggregation on MSR-VTT dataset. The last row is our design. }
\label{table:ab1}
\end{table}

\subsection{Results with Full Metrics}
In this section, we report the video-text retrieval results on MSR-VTT \cite{xu2016msr}, ActivityNet \cite{krishna2017dense} and DiDeMo \cite{anne2017localizing} with full video-text retrieval metrics, including the results on video-to-text retrieval setting. 

\noindent \textbf{MSR-VTT.} 
As shown in \Cref{table:msrvtt}, \Modelname{} achieves state-of-the-art results on most metrics. Specifically, compared to the most recent multi-level alignment method X-CLIP \cite{ma2022x}, \Modelname{} achieves a $3.3\%$ gain on text-to-video R@1 metric and obtains comparable results on video-to-text retrieval metrics. Compared to another recent state-of-the-art CLIP-based method TS2-Net \cite{liu2022ts2}, our model gets $2.4\%$ and $1.8\%$ improvement on R@1 metric for text-to-video and video-to-text retrieval. These results verify the effectiveness of the \Modelname{} model. Moreover, replacing the visual backbone (ViT-$32$) with a larger model (ViT-$16$) would improve the model performance, especially on video-to-text retrieval.

\noindent \textbf{ActivityNet.} 
As shown in \Cref{table:anet}, \Modelname{} outperforms the current state-of-the-art CLIP-based methods on a wide range of metrics on ActivityNet benchmark \cite{krishna2017dense}. Concretely,  our model achieves $1.4\%$ and $2.4\%$ gain on the R@1 metric on text-to-video and video-to-text retrieval compared to the state-of-the-art approaches. This indicates our \Modelname{} model is capable of tackling long video retrieval, thus validating the generalization ability of our method.

\noindent \textbf{DiDeMo.} 
As shown in \Cref{table:didemo}, compared to the current state-of-the-art models, \Modelname{} achieves better results on most evaluation metrics. Specifically, our model outperforms the recent state-of-the-art CLIP-based approach X-CLIP \cite{ma2022x} with a significant margin of $1.3\%$ on text-to-video R@1 and $2.9\%$ on video-to-text R@1.

\begin{figure}[t]
\begin{center}
   \includegraphics[width=1\linewidth]{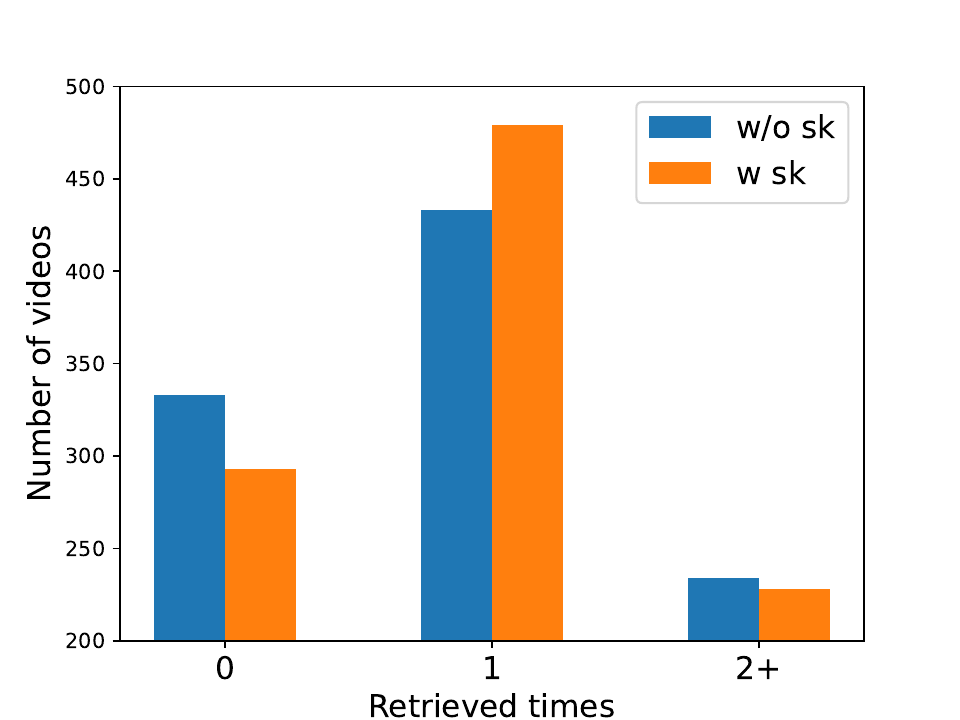}
\end{center}
   \caption{The visualization of imbalanced retrieval results. The left part denotes the video candidates haven't been retrieved in the inference stage (under-representative). The middle part denotes the video candidates have been retrieved once in the inference stage. The right part denotes the video candidates have been retrieved more than twice (including twice) in the inference stage (over-representative). The blue column represents the model without the Sinkhorn Knopp algorithm and the orange column represents the full \Modelname{} model. Results show that about $50$ under-represented videos have been re-scaled and retrieved by the videos via the SK algorithm (from the left bar to the middle bar). }
\label{fig:vis1}
\end{figure}

\begin{figure*}[t]
\begin{center}

   \includegraphics[width=1\linewidth]{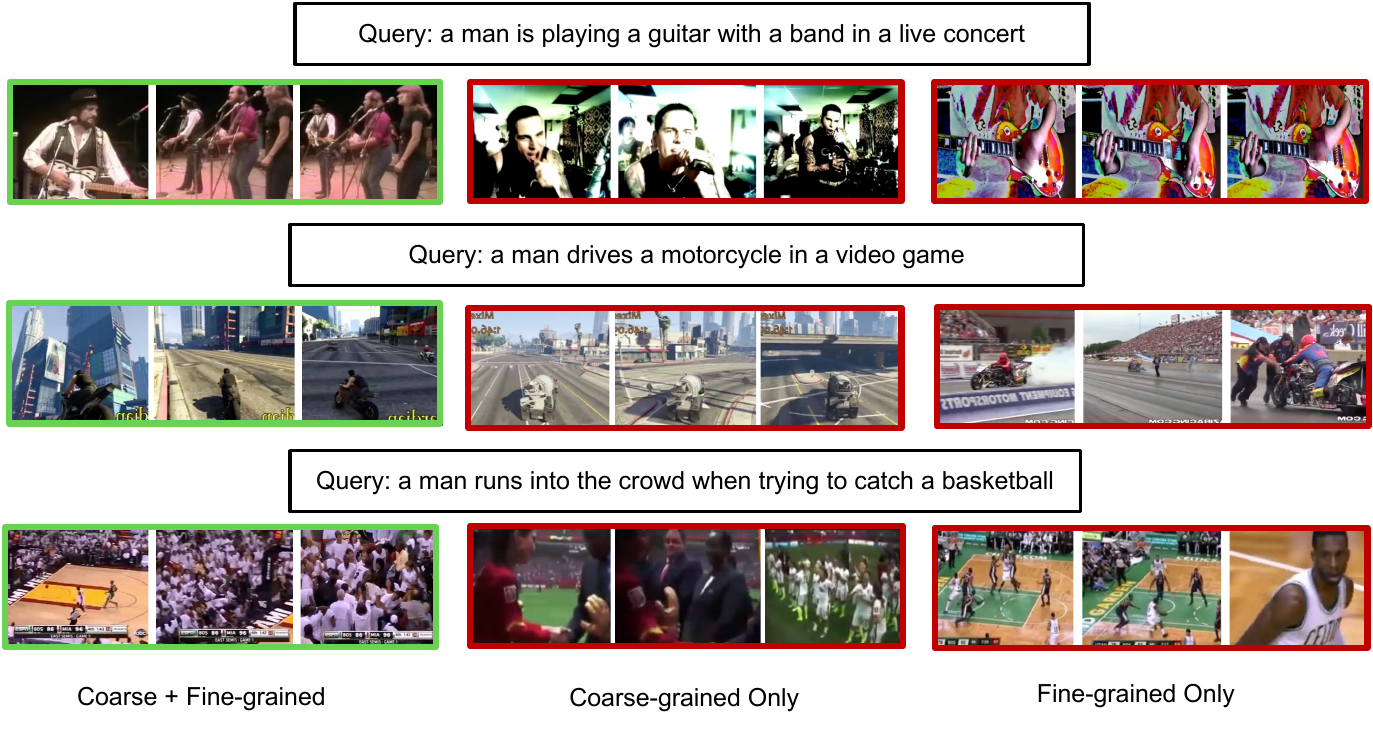}
\end{center}
   \caption{The visualization of different alignments. The left part is the correctly retrieved video by our coarse-to-fine alignment module. The middle part is the wrongly retrieved video by coarse-grained only alignment and the right part is the wrongly retrieved video by fine-grained only alignment. }
\label{fig:vis_2}
\end{figure*}

\subsection{Adapt \Modelname{} to Other Backbone Model}
In this section, we apply \Modelname{} to the recent CLIP-ViP's \cite{xue2022clip} backbone model, which is a video-text model pretrained on $100$M video-text pairs. As shown in \Cref{table:vip}, \Modelname{} improves the backbone CLIP-ViP model on all metrics on the MSR-VTT text-to-video retrieval task. This indicates that our method is able to generalize to a more advanced backbone model and verifies the robustness of our method.

\subsection{The Computational Cost of \Modelname{}}
In this section, we compare our model with the recent X-CLIP model \cite{ma2022x} on the balance of model performance and computational cost in \Cref{table:computation}. Results show that \Modelname{} is $\textbf{3.3\%}$ better than X-CLIP on text-to-video retrieval on MSR-VTT dataset while only requiring $\textbf{1.2\%}$ additional parameters and $\textbf{1.4}$ GB memory per GPU (train on $4$ GPUs). Therefore, \Modelname{} achieves significant improvement with limited additional computational cost compared to previous works. 

\subsection{Different Training Settings for Encoders}
In this section, we show the performance of \Modelname{} under different training settings for encoders. Our approach attains $47.1\%$, $49.4\%$, $43.8\%$ R@1 on text-to-video retrieval on MSR-VTT dataset when using $1\mathrm{e}{-6}$, $1\mathrm{e}{-7}$ and $0$ (frozen) learning rates for the encoders, respectively. We conjecture that CLIP is pretrained on very large-scale image-text pairs (400M) and thus we only need to slightly adjust its parameters for downstream tasks, which is also observed by many previous works (CLIP4Clip, X-CLIP). 

\subsection{Additional Ablation Study for Bi-ISA}
In the main paper, we mention that empirically we find that jointly considering two directions of patch-word matrix aggregation (patch-then-word and word-then-patch) provides better aggregation for the patch-word matrix. In \Cref{table:ab1}, we compare our bi-directional solution with single-directional methods on the MSR-VTT dataset. For better comparison, we do not apply the Sinkhorn Knopp algorithm to normalize the retrieval similarities. Results show that leveraging both aggregation directions achieves better results, validating the effectiveness of our Bi-ISA design.

\section{Additional Qualitative Results}

In this section, we provide additional qualitative results of \Modelname{}. First, we visualize the imbalanced retrieval results and show how our unification module mitigates this issue. Then, we visualize the video samples retrieved by methods focusing on different alignment levels to validate the effectiveness of our coarse-to-fine alignment design. 

\subsection{Visualization of Imbalanced Retrieval}

As discussed in the main paper, we find that scores across different videos are highly imbalanced in the similarity matrices of each level. As a result, the video candidate could be over-/under-represented by the retrieval model due to the imbalanced summation of retrieval similarities. As shown in \Cref{fig:vis1}, the left part denotes the video candidates haven't been retrieved in the inference stage which corresponds to under-representative. The right part denotes the video candidates have been retrieved more than twice (including twice) in the inference stage which corresponds to over-representative. The middle part denotes the video candidates have been retrieved once, which is the ideal situation. The blue column in \Cref{fig:vis1} represents the model without the Sinkhorn Knopp algorithm. The results show that only $43\%$ video candidates are retrieved once in the inference stage while $34\%$ video candidates are under-represented and $23\%$ video candidates are over-represented. After applying the Sinkhorn Knopp algorithm in the unification module (the orange column in \Cref{fig:vis1}), the under-representative issue is mitigated and more than $50$ under-represented video candidates have been re-scaled and retrieved by the model. Meanwhile, we also observe a slight reduction in the number of over-represented videos. In all, the Sinkhorn Knopp algorithm in the unification module indeed mitigates the over- and under-representation issue in the inference stage. 

\subsection{Comparison of Different Alignments}
As discussed in the main paper, our coarse-to-fine alignment module captures comprehensive cross-modal clues compared to coarse-grained or fine-grained alignment. We provide more visualization results in \Cref{fig:vis_2}. For the first text query (on the first row of \Cref{fig:vis_2}), the coarse-grained alignment only captures the scene of ``singing'' and the fine-grained alignment only focuses on the object ``guitar''. For the second text query (on the second row of \Cref{fig:vis_2}), the coarse-grained alignment only considers the scene information like ``driving'', and ``video game'' while the fine-grained alignment only captures the detail information ``motorcycle''. For the last text query (on the last row of \Cref{fig:vis_2}), the coarse-grained alignment overlooks the detailed information ``basketball'' and the fine-grained alignment ignores the scene of ``crowd'' and the action of ``run into''. To sum up, the coarse-grained or fine-grained alignment could overlook some crucial cross-modal clues while our coarse-to-fine alignment is capable of capturing both high-level and detailed information and retrieving the correct video candidate.

\section{Method Details}

In this section, we present more details of \Modelname{}. First, we discuss the patch selection module. Then, we present details of the Sinkhorn-Knopp Algorithm that normalizes the similarity matrix for unification. 

\subsection{Patch Selection Module} 

As discussed in the main paper, due to the high redundancy of patch tokens, inspired by \cite{liu2022ts2}, we propose a patch selection module to choose the top-K salient patches from each frame for patch-word alignment. Here we present the details of the patch selection module. 

Specifically, given the patch feature for the $n$-th frame ${p}_n$, where ${p}_n = \mathcal{F}_v \left( F_n \right)  \in \mathbb{R}^{M  \times C} $, $M$ denotes the number of the visual patches within a video, we select the top-$K$ salient token out of the $M$ tokens of the frame. To allow each patch to be aware of the information of the whole frame, we first concatenate the frame feature $f \in \mathbb{R}^{C}$ with each patch feature and leverage an MLP layer to fuse the global (frame) and local (patch) information and leverage an MLP layer $\mathcal{G}_a$ to obtain the frame-augmented patch information to mitigate the influence of irrelevant background patches. Then, to avoid the selection module only considering the frame information and deviating from the information of the original video, we further concatenate the frame-augmented patch information with the video representation $v$ and apply another MLP layer $\mathcal{G}_b$ to obtain a saliency score $U$ for each patch. The whole process can be denoted as:

\begin{equation}
U = \mathcal{G}_b \left( \operatorname{Concat}\left( \mathcal{G}_a \left( \operatorname{Concat}\left( {p}_n , f \right) \right) , v \right)\right).
\label{eq:token selection}
\end{equation}

Then, according to the saliency score $U$, we select the indices of $K$ most salient patches within a video frame $ind \in \{ 0,1  \} ^{K}$. Through this one-hot vector $ind$, we extract the top-K salient patch by 
\begin{equation}
\hat{p} = ind^T p,
\label{eq:token mul}
\end{equation}
 where $\hat{p}_n \in \mathbb{R}^{K \times C}$ denotes the selected patch representation for the whole video. We concatenate the selected patch feature from all $N$ frames and obtain the selected patch feature $\hat{p} \in \mathbb{R}^{L_v \times C}$, where $L_v = N*K$. Note that the direct top-K patch selection is non-differentiable, in practice,  to make the patch selection module differentiable, we apply the perturbed maximum method proposed in \cite{berthet2020learning}.

\begin{algorithm}[t]
\caption{Sinkhorn-Knopp algorithm}\label{alg:cap}
\begin{algorithmic}

\Function{Sinkhorn-Knopp}{$\textbf{S}, n_{iter}$}

\State $L = \textbf{S}.\exp()$
\State $\beta = 1  \, / \, L.\text{sum}(\texttt{dim} = 0)$

\For{$i$ $  \textbf{in range} (n_{iter})$}  
    \State $\alpha = 1  \, / \, ( L  \, \operatorname{@} \,  \beta )$
    \State $\beta = 1  \, / \, ( \alpha \, \operatorname{@} \, L )$
\EndFor

\State $\alpha \leftarrow \alpha.\log()$
\State \textbf{return} $\alpha$

\EndFunction
\end{algorithmic}
\end{algorithm}

\subsection{Sinkhorn-Knopp Algorithm}

As discussed in the main paper, inspired by \cite{park2022normalized}, we utilize the Sinkhorn-Knopp algorithm \cite{cuturi2013sinkhorn} to normalize the similarity scores for each granularity and make sure the marginal similarities (the sum of retrieval similarities between one specific video and all texts) for different videos are almost identical so that each video has a fair chance to be selected. Below, we discuss the algorithm in detail. 

Recall that our goal is to compute the video bias using the testing video set ($G$ videos) and the training text set ($H$ queries). Given the similarity matrix $\textbf{S} \in \mathbb{R}^{G \times H}$, we leverage the \Cref{alg:cap} to compute the video bias $\alpha \in \mathbb{R}^{G}$ in an iterative manner (the number of iterations $n_{iter} = 4$ for all datasets). The fixed-point iteration process allows the model to find the optimal value of $\alpha$ with minimum cost. We further add the $\alpha$ to the similarity logits to re-scale the similarity matrix to normalize the marginal similarity of every video to be a similar value.

\end{document}